\pdfoutput=1
\documentclass[11pt]{article}
\usepackage{amsmath}
\usepackage[final]{coling}

\usepackage{times}
\usepackage{latexsym}
\usepackage[T1]{fontenc}
\usepackage[utf8]{inputenc}
\usepackage{microtype}
\usepackage{inconsolata}
\usepackage{graphicx}
\usepackage{multirow}

\usepackage{tikz-dependency}
\usepackage{booktabs}
\usepackage{tabularx}
\usepackage{adjustbox}
\usepackage{arydshln}
\usepackage{subcaption}
\usepackage{xspace}
\usepackage{pifont}
\usepackage{makecell}
\usepackage{longtable}

\usepackage{tipa}
\usepackage{comment}

\newcommand*{\gloss}[1]{\textcolor{gray}{\textit{#1}}}

\newcommand{\punit}[6]{%
& #1 & #4\\
& #2 & #5\\
& #3 & \gloss{#6}\\[3pt]%
}

\title{Neural Text Normalization for Luxembourgish using \\ Real-Life Variation Data} 

\author{
 \textbf{Anne-Marie Lutgen\textsuperscript{1}},
 \textbf{Alistair Plum\textsuperscript{1}},
 \textbf{Christoph Purschke\textsuperscript{1}},
 \textbf{Barbara Plank\textsuperscript{2,3}},
\\
\\
 \textsuperscript{1}University of Luxembourg, Esch-sur-Alzette, Luxembourg \\
 \textsuperscript{2}MaiNLP, Center for Information and Language Processing, LMU Munich, Germany, \\
 \textsuperscript{3}Munich Center for Machine Learning (MCML), Munich, Germany,
\\
 \small{
   \textbf{Correspondence:} \href{mailto:anne-marie.lutgen@uni.lu}{anne-marie.lutgen@uni.lu}
 }
}

\begin{document}
\maketitle
\begin{abstract}
Orthographic variation is very common in Luxembourgish texts due to the absence of a fully-fledged standard variety. Additionally, developing NLP tools for Luxembourgish is a difficult task given the lack of annotated and parallel data, which is exacerbated by ongoing standardization. In this paper, we propose the first sequence-to-sequence normalization models using the ByT5 and mT5 architectures with training data obtained from word-level real-life variation data. We perform a fine-grained, linguistically-motivated evaluation to test byte-based, word-based and pipeline-based models for their strengths and weaknesses in text normalization. We show that our sequence model using real-life variation data is an effective approach for tailor-made normalization in Luxembourgish.

\end{abstract}

\section{Introduction}
Automatic text normalization is the task of mapping non-standard spellings to a standard \cite{han-baldwin-2011-lexical,van_der_goot_monoise_2019}. Normalization thus reduces orthographic variation and noise in language data. It can serve as a pre-processing step to facilitate downstream tasks like POS-tagging and NER~\cite[e.g.][]{kucuk-steinberger-2014-experiments,van-der-goot-cetinoglu-2021-lexical}.

In this paper, we address orthographic normalization for Luxembourgish, a Germanic language currently in the process of political development, including orthographic standardization \cite{Gilles_uberdachung}. Spelling norms for Luxembourgish are not a novelty, however, due to a lack of language teaching in school, written Luxembourgish today is characterized by vast amounts of variation, e.g., orthographic, lexical, syntactical or regional.
This has led to written Luxembourgish texts adhering to the standard orthography to be rare, even in formal contexts. For this reason, we develop an automatic text normalization model for Luxembourgish to reduce variation in written data as a pre-processing step for NLP tasks.
 
Luxembourgish is an under-researched language, and as such there is a lack of annotated parallel data for training and fine-tuning normalization models.
To tackle this problem, our proposed solution uses word-level real-life variation data to create training data sequences and fine-tune multilingual sequence-to-sequence models. In this paper, we use ByT5 \cite{byt5} and mT5 \cite{mt5} models and in addition, we benchmark the generative models GPT-4o and Llama and a word-based Luxembourgish correction pipeline, \textit{spellux}.\footnote{\url{https://github.com/questoph/spellux}} 

We evaluate model performance using both quantitative metrics and a tailored qualitative evaluation of linguistic contexts.
Then, we compare byte-based, word-based and pipeline-based models to identify linguistic contexts in which models perform particularly well or struggle.

\begin{figure*}[ht!]
\centering
\includegraphics[scale=0.15]{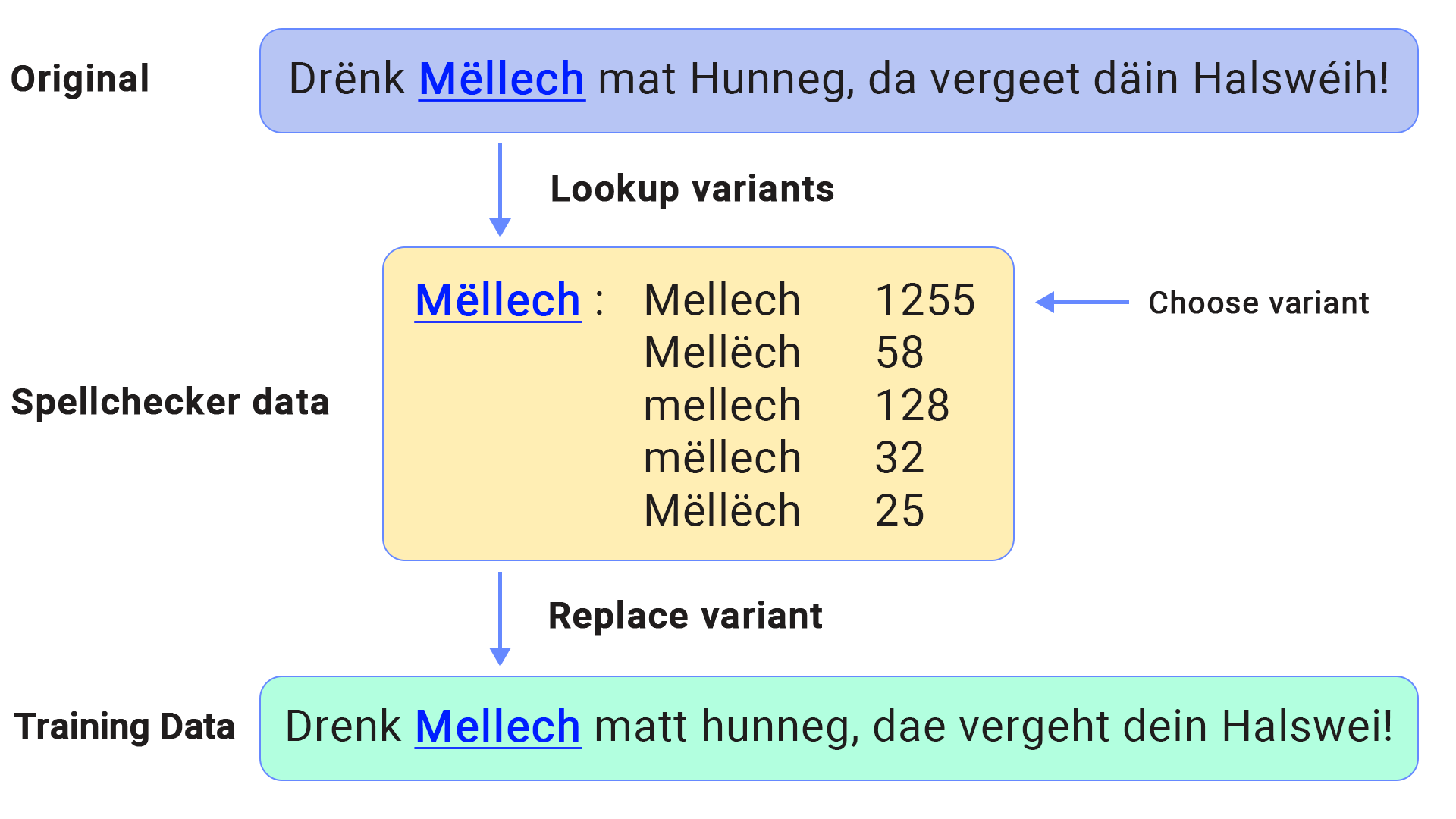}
\caption{Illustration of the creation of training data with the Luxembourgish Online Dictionary (LOD) sentence `Drink milk with honey, then your throat will no longer hurt.' and the variational statistical data for `milk'. The algorithm processes every word sequentially, this illustrates only the replacement process for the word `milk'. }
\label{fig:synth}
\end{figure*}

Our main contributions are therefore twofold: 
\begin{itemize}
    \item[(1)] The first generative normalization model for Luxembourgish, trained on real-life variation data obtained from an online spellchecker.
    \item[(2)] A linguistically-informed qualitative test set tailored for Luxembourgish orthography, besides a comprehensive quantitative evaluation. 
\end{itemize}

\section{Related Work}
Luxembourgish, a relatively small language, is less represented in NLP compared to its linguistic neighbors, French and German. Research in NLP for Luxembourgish has only recently gained momentum, with a few earlier works: \citet{adda-decker-etal-2008-developments} introduced various resources for NLP tasks in Luxembourgish; \citet{snoeren-etal-2010-study} analyzed typical writing patterns (contextual n-deletion) in transcribed speech; and \citet{lavergne-etal-2014-automatic} presented a manually annotated corpus of mixed-language sentences to test a word-based language identification system. Additionally, the first treebank for Luxembourgish \textit{Luxbank} was recently released \cite{plum2024luxbank}.

Other developments include \citet{sirajzade-etal-2020-sentiment} and \citet{Gierschek2022}, who tested various approaches for performing sentiment analysis on Luxembourgish, including training BERT-based models. \citet{philippy-etal-2024-forget} proposed a new approach to zero-shot classification using a task-specific dictionary for topic classification. For spoken data, \citet{gilles-etal-2023-asrlux} and \citet{gilles-etal-2023-luxasr} developed \textsc{LUX-ASR}, an efficient Automatic Speech Recognition system for Luxembourgish. Additionally, \citet{ranasinghe-etal-2023-publish} fine-tuned language models for automatic comment moderation. Some language models have been trained using transfer learning from German, such as \textsc{LuxGPT} \cite{bernardy2022}. Other models developed for Luxembourgish are \textsc{LuxemBERT} \cite{lothritz-etal-2022-luxembert} and \textsc{ENRICH4ALL} \cite{anastasiou-2022-enrich4all}. 

Various T5 \cite{t5} and ByT5 \cite{byt5} architectures have been developed for lexical normalization. \citet{ufal} pre-trained a ByT5 model for 12 languages with synthetic data as part of the shared task MultiLexNorm \cite{multilexnorm}. Similarly, \citet{rothe} fine-tuned a mT5 model using synthetic parallel data for German, English, Czech and Russian. \citet{kuparinen_dialect--standard_2023} evaluated different sequence-to-sequence models including ByT5 for dialect-to-standard normalization in Norwegian, Swiss German, Slovene and Finnish.  

\citet{lusetti_encoder-decoder_2018} developed an encoder-decoder architecture for text normalization in Swiss German by using sequence-to-sequence models but not T5 architectures. Similarly, \citet{bollmann_normalization_nodate} worked on historical text normalization, comparing encoder-decoder architectures to statistical machine translations. 

For pipeline-based normalization, \citet{van_der_goot_monoise_2019} developed MoNoise, which has long been regarded as the state-of-the-art text normalization tool. This pipeline operates at a word level using a spelling correction module and word embeddings for various languages. Furthermore, \citet{van_der_goot_monoise_2019} introduced the error reduction rate (ERR) as an evaluation metric for normalizers. MoNoise has also been used on the task of nested named entities in Danish \cite{plank_dan_2020} and for code-switching data \cite{van-der-goot-cetinoglu-2021-lexical}. For Luxembourgish, \citet{purschke2020attitudes} published \textit{spellux} a pipeline for automatic orthographic correction of text data, the first text normalization tool for Luxembourgish. 

\section{Methodology}
This section describes the methodology for fine-tuning and evaluating our normalization model for Luxembourgish. This includes the creation of training data, the experimental setup for the model training and benchmarking, and the model evaluation process. 

\subsection{Creating Training Data}
The lack of annotated and parallel datasets in Luxembourgish is a challenge for developing tailored NLP solutions. This not only applies to text normalization but also, for example, to NER and machine translation tasks. Creating parallel datasets manually or via crowdsourcing is not a viable option for Luxembourgish, as the majority of the population has no formal training in orthography due to the lack of extensive grammar teaching in school contexts. Luxembourgish has only recently been integrated more into the school curriculum to foster the societal anchoring of the spelling rules. As a consequence, corpus data for Luxembourgish written in the standard orthography are scarce. The solution applied in this paper is to create training data based on real-life variation data obtained from an online spellchecking tool.

Creating training data from synthetic data for normalization has been used on multiple occasions, as in \citet{ufal} and \citet{rothe}. However, using only synthetic data may be problematic as it does not accurately represent real-life language use. For this paper, we use data provided by the spelling correction website \textit{Spellchecker.lu}\footnote{\url{https://spellchecker.lu}} as a basis for the creation of training data. Users on the website can manually correct written Luxembourgish text based on context-sensitive suggestions offered by the system. Pairs of entered and corrected words are then logged and statistically aggregated. As a consequence, this dataset offers a unique real-life dictionary of spelling variants per lemma, including their frequency of use.

Using real-life variation to create training data ensures that each variant in the data has actually been used by people and is not just a random character replacement. Additionally, the frequency of use of these variants can be represented in the data realistically. Since the baseline for the replacements mirrors a realistic distribution of spelling variants based on actual texts written by people, this approach is considered superior to using synthetic data. The training data, hence, captures the actual patterns of variation in Luxembourgish and is not a randomly assembled approximation of non-standard texts. 

As the website is widely known and used in the country,\footnote{The popularity of the website can be explained by the recent increase in the use of written Luxembourgish in the population, without having formal orthography training. The website provides a helpful tool for writing Luxembourgish correctly, e.g., in formal contexts.} Spellchecker.lu provides us with an extensive overview of the orthographic variation space in Luxembourgish. The variant dictionary includes 138,802 different lemmas with numerous variants per lemma. Figure \ref{fig:synth} illustrates an example with the word \textit{Mëllech} (`milk') and its most frequent variations in the Spellchecker data. 

We use transcriptions of discussions in the Chamber of Deputies in Luxembourg as a source of orthographically correct Luxembourgish for the training data.\footnote{\url{https://www.chd.lu/de/chamberblietchen}} These transcriptions are from 2002-2012 and 2019-2020 and are produced by trained writers, ensuring the correctness of the texts.

Combining the Spellchecker.lu variant dictionary and the transcriptions then allows for the creation of parallel training data using correct Luxembourgish texts and real-life variation patterns. We apply an algorithm that processes every word in the original sentence sequentially and looks up the variants in the Spellchecker.lu data. If the lemma is part of the variant dictionary, it is replaced by a variant based on its frequency of use. This process results in around 833,000 parallel standard/non-standard sentence pairs with a mean token difference of 19 tokens being changed per sentence pairs that can be used as training data. Figure \ref{fig:synth} illustrates this process with an example sentence taken from the Luxembourgish Online Dictionary (LOD).\footnote{\url{https://lod.lu}}

\subsection{Experimental Setup}
We fine-tune two multilingual sequence-to-sequence models, ByT5 and mT5. For benchmarking the task, we prompt Llama\footnote{
Llama-3.1-8B-Instruct} and GPT\footnote{gpt-4o-2024-08-06}, as well as using the word-based normalization tool \textit{spellux}. In this way, we test various approaches for Luxembourgish text normalization, i.e., a byte-based sequence-to-sequence and word-based sequence-to-sequence method, as well as generative models and a word-based pipeline. Due to a lack of training data we did not opt for pre-training a sequence-to-sequence model ourselves for this task. However, recently \citet{plum2024textgenerationmodelsluxembourgish} pre-trained a T5-based model with multilingual data to improve performance for Luxembourgish.  

ByT5 is a multilingual byte-based sequence-to-sequence model which encodes the input sequence to UTF-8-encoded bytes and produces an output sequence of UTF-8-encoded bytes \cite{byt5}. The robustness to noise and variation of byte-based and character-based models \cite{byt5} makes them ideal models to fine-tune for normalization in Luxembourgish. mT5 is a multilingual transformer encoder-decoder model trained on 101 languages \cite{mt5}. While ByT5 is the focus of our experiment, the word-based model mT5 is used as a comparison to the byte-based approach for a sequence-to-sequence task.

The fine-tuning setup stays the same for the ByT5 and mT5 models. Our experiments focus on testing the experimental method (real-life training data and comprehensive performance testing) rather than producing an optimized model, although we did perform some hyperparameter tuning, where we were not constrained by hardware limitations.
Using the ByT5 base model with 582M parameters, the best performing model has a batch size of 16, a learning rate of 1e-4,  and a sequence length of 256 trained on 3 epochs. We also fine-tune the ByT5 large model with 1.23B parameters to test the influence of parameter size on task performance. We restrict the hyperparameter setup for fine-tuning to a sequence length of 128 and an epoch number of 1. The mT5 model is fine-tuned using the base variant with 582M parameters. Additionally, we fine-tune one mT5 model with the same hyperparameters as the ByT5 model, a sequence length of 128, 1 epoch and batch size 2. 

Benchmarking is done by prompting GPT-4o and Llama 3.1. The setup is the same for both models, using the following prompt: ``You are a Luxembourgish teacher. Your task is to correct these sentences on a word level based on the correct Luxembourgish orthography. Please only write the corrected sentence and no explanation''. For this task, we use the same evaluation sentences as for the other models.
The main focus of the setup lies on models that are developed specifically for Luxembourgish, nonetheless we include GPT-4o and Llama for completeness reasons as the results are not reproducible and there is a lack of knowledge concerning the training data (see Section \ref{sec:disc}). 

Our main comparison of the models is with the \textit{spellux} text correction pipeline, which is developed specifically for Luxembourgish. The tool implements a combination of correction algorithms for candidate evaluation: a word-based embedding model trained on the entire archive from RTL.lu (journalistic texts and user comments), an adapted version of the spelling correction tool written by Peter Norvig\footnote{\url{https://norvig.com/spell-correct.html}}, and an ngram-based tf-idf similarity matrix based on the RTL corpus. \textit{spellux} also includes an adapted version of the variant dictionary created from the Spellchecker data. For benchmarking, we use the default settings of the pipeline. 

\begin{table*}[htp]
\begin{adjustbox}{max width=\textwidth, center}
\begin{tabular}{@{}lll@{}}
\toprule
\multicolumn{2}{l}{Category} & Test Sentence \\ \midrule
\multicolumn{2}{l}{\bf Quantity rule}\\
\hspace{0.2em}
\punit
{writing of long vowels depending on stressed vowels \& consonants}
{\cite{peter_codification}}
{}
{Wou ass d'\underline{Bischt} fir ze kieren?}
{\textit{Correction:} Biischt}
{Where is the broom to sweep (with)?}

\multicolumn{2}{l}{\bf Short Vowels}\\
\hspace{0.2em}
\punit
{stressed short vowels and consonants}
{\cite{peter_luxembourgish}}
{}
{D'Haus ass op mech \underline{geschriwen}.}
{\textit{Correction:} geschriwwen}
{The house is written under my name.}

\bottomrule
\end{tabular}%
\end{adjustbox}
\caption{%
Selection of test units for Luxembourgish. Full set of rules with examples provided in the Appendix.}
\label{tab:performance-ex}
\end{table*}
\renewcommand{\arraystretch}{1}

\subsection{Evaluation}\label{sec:evaluation} 
We perform a comprehensive evaluation of the models based on both quantitative metrics and a qualitative analysis, where we compare the output of different models to gain more insight into how well the different models solved the task. First, we perform a quantitative evaluation using a wide array of evaluation metrics, then, we develop a set of qualitative tests tailored to the normalization task, inspired by CheckList \cite{beh_tests}. This allows for a linguistically informed and systematic analysis of the output and performance of the used models. 

\paragraph{Quantitative} For the quantitative evaluation, we create a corpus consisting of random user comments from the RTL media platform, as they contain a high amount of variation \cite{purschke2020attitudes}, and correct them manually. This results in an evaluation corpus consisting of 459 sentences from the comments, equalling 7,146 tokens.  

The evaluation metrics used for the fine-tuned models and benchmarking include accuracy, recall, precision, F1-score, and error reduction rate (ERR) at the word-level, and character error rate (CER) at the character level. 
To calculate the word-level metrics, we align the original sentences, the predicted output sentences and the orthographically correct sentences at the word-level. 
The alignment is done by repurposing the \emph{Needleman-Wunsch} algorithm \cite{needleman_wunsch} with Kamil Slowikowski's code for 3D alignment and string alignment\footnote{\url{https://gist.github.com/slowkow/06c6dba9180d013dfd82bec217d22eb5}}, using the Levenshtein distance for fuzzy string matching. Although other distance metrics are available, we did not carry out any experiments with these as this was not within the scope of our research. We therefore opted for the Levenshtein distance, since it is well established.

The most important metric for the normalization task is the ERR, introduced by \citet{van_der_goot_monoise_2019} as an evaluation metric for normalizers, and used as the main evaluation metric in \citet{multilexnorm}. It captures the accuracy normalized over the number of words to be corrected \cite{van_der_goot_thesis}. The ERR normally has a value between 0 and 1. Zero represents the leave-as-is baseline, a negative value indicates that the model performs worse than the baseline, and a positive value means that the model normalizes more words correctly. The comparability across multiple corpora is the main advantage of using this metric, as the ERR is a normalized value \cite{van_der_goot_thesis}. 

Besides word-level metrics, the character-level metric CER is included so that the evaluation becomes more granular. This means being able to not only distinguish between words that are either simply correct or incorrect, but also by how many characters words have changed \cite{kuparinen_dialect--standard_2023}. While this is by no means an indicator for degrees of correctness, the metric does allow for gauging how far away a predicted sentence is from its correct form.\footnote{The CER is calculated using the implementation available at \url{https://github.com/nsmartinez/WERpp} following \citet{kuparinen_dialect--standard_2023}} 

\paragraph{Qualitative} For the qualitative evaluation, we use a setup similar to CheckList \cite{beh_tests}, a methodology to systematically test NLP models, to evaluate the performance of the normalizer. Specifically, we use the Minimum Functionality test to probe the model as to the handling of Luxembourgish orthographic rules and to gain more linguistic insights into the strengths and weaknesses of the different models. These tests include two different setups and implement 21 orthographic rules. These rules are implemented based on the official Luxembourgish orthography.\footnote{ {\href{https://portal.education.lu/Portals/79/Documents/WEB_LetzOrtho_Oplo5_v02-1.pdf?ver=2021-01-13-085421-963}{D’Lëtzebuerger Orthografie, ZLS 2022.}}} The first setup tests the traditional application of a normalizer by correcting an incorrect target word, therefore checking corrections systematically against the backdrop of orthographic rules. The target word is corrupted systematically by applying the orthographic rule in reverse. 

The second setup tests false positives by giving a correct input and examining the number of false corrections proposed by the model. We include this test because of the known issues with automatic text normalization, which might increase the number of incorrect forms in a given text. This is also captured in the ERR, as a value under 0 indicates more mistakes than before. We include 10 sentences per test setup per category, which results in 420 sentences.\footnote{All sentences are taken from the \href{https://lod.lu}{LOD}.}

Table \ref{tab:performance-ex} shows selected categories, with a short description and a test sentence from the first setup. Appendix A includes the full table with the 21 rules following the same format. The tables also include references to linguistic literature for each respective phenomenon. The first category in Table \ref{tab:performance-ex} is the \textit{quantity rule} which describes the use of the long vowels <a, i, o, u, ä, ö, ü>. The test sentence stems from the first setup and the underlined word \textit{Bischt} (`broom') is the target word, that the model should correct into the correct form \textit{Biischt}. 

\begin{table*}[ht!]
\centering
\scalebox{0.9}{
\begin{tabular}{|l||c|c|c|c|c|c|c|c|}
\hline
\textbf{Model}    & \textbf{Accuracy} & \textbf{Recall} & \textbf{Precision} & \textbf{F1-Score} & \textbf{ERR} & \textbf{CER} \\ \hline \hline
ByT5 base        & 78.8    & 54.8          & 65.9             & 59.8             & \textbf{0.26}         & 11.7        \\ \hline
ByT5 large        & 71.8     & 51.3           & 49.6              & 50.4              & -0.01       & 20.4         \\ \hline
mT5               & 27.6     & 35.5           & 5.7               & 9.7               & -5.70         & 22.2         \\ \hline
Llama             & 63.7     & 0.0            & 0.0               & 0.0               & -0.15        & 10.7         \\ \hline
GPT-4o           & 84.8     & 66.0           & 77.5              & 71.3              & \textbf{0.46}         & 7.2          \\ \hline
spellux           & 82.2     & 46.8           & 86.3              & 60.7              & \textbf{0.39}         & 7.5          \\ \hline
\end{tabular}
}
\caption{Evaluation of models, scores are in percentages except ERR.}
\label{tab:model-performance}
\end{table*}

\section{Results}
This section illustrates the results from the comprehensive quantitative evaluation and the linguistically-informed qualitative tests. The evaluated models include fine-tuned ByT5 and mT5 models, generative models GPT and Llama and the pipeline-based \textit{spellux}.  

\subsection{Quantitative}
Table \ref{tab:model-performance} shows all the models trained on the normalization task for Luxembourgish, including the benchmarking with GPT, Llama and \textit{spellux}. It is a comparison between a byte-based, word-based, generative-based and pipeline-based normalization method for Luxembourgish. As already established, the ERR is the most important metric for normalization. 

The ByT5 base model is the best performing model using T5 architecture for Luxembourgish, with an ERR of 0.26, an accuracy of 78.79\% and a precision of 65.9\%, taking into account that this model is pre-trained on multilingual data. In comparison, ByT5 large, for which we did not perform any hyperparameter optimization, only reproduces the leave-as-is baseline with an ERR of -0.01. In contrast, mT5 performs the worst among the T5 architectures. Accuracy and precision are very low, as is the ERR. Additionally, the CER is the lowest for the ByT5 base model, indicating fewer mistakes in a corrected corpus than in other models.
Hence, the byte-based model is more suitable for Luxembourgish text normalization than the other tested models. 

Recurring issues with both ByT5 models are hallucination, including the repetition of training data and stopping early with long sentences. \citet{kuparinen_dialect--standard_2023} encountered similar issues with stopping early. However, an increase in epochs and sequence length when training the ByT5 base model reduces the hallucination rate to 5\% and the stopping rate to 2\%.

The benchmarked generative models perform very differently from each other. Llama shows an even worse performance than mT5 with an ERR of -0.15. The accuracy of 63.7\% is not much lower than the other models, but Llama achieves 0 true positives and therefore a F1-score of 0. In comparison, GPT-4o performs well, with the highest ERR score for this task. An important factor to consider is the rapid progress of GPT and therefore the issue of reproducibility with these generative models. The benchmarking results using the 3 month older gpt-4o-2024-05-13 are much lower than the current results with an ERR of 0.12. This demonstrates how quickly GPT has improved, albeit with a lack of transparency.

In contrast, the pipeline-based model \textit{spellux} has a good performance overall. In particular, the high ERR rate of 0.39 indicates a high correction rate. Only recall is lower than for the ByT5 base model. 

\begin{table*}[ht!]
\centering
\scalebox{0.9}{
\begin{tabular}{|l|cc|cc|cc|}
\hline
\multirow{2}{*}{\textbf{Category}} & \multicolumn{2}{c|}{\textbf{ByT5 base}} & \multicolumn{2}{c|}{\textbf{mT5}} & \multicolumn{2}{c|}{\textbf{spellux}} \\
                  & \textbf{correct} & \textbf{preserve} & \textbf{correct} & \textbf{preserve} & \textbf{correct} & \textbf{preserve} \\ \hline
Quantity Rule              & 80 & 80    & 20  & 100 & 80 & 100 \\ \hline
Short Vowels                & 70    & 100   & 20  & 100 & 50 & 100 \\ \hline
Short Open Vowel [\textipa{\ae}]               & 50    & 100   & 10  & 100 & 50 & 100 \\ \hline
Short Closed Vowel [\textipa{e}]             & 70    & 90    & 20  & 100 & 40 & 100 \\ \hline
Neutral Short Vowel [\textipa{@}]            & 90    & 100   & \textbf{40}  & 100 & 70 & 100 \\ \hline
Long Vowel [\textipa{e:}]               & 40    & 100   & 0   & 100 & 30 & 100 \\ \hline
Diphthongs                 & 60    & 100   & 20  & 100 & 50 & 100 \\ \hline
r-Rule                     & 70    & 100   & 0   & 100 & 60 & 100 \\ \hline
Final Devoicing            & 60    & 100   & 30  & 90  & 40 & 100 \\ \hline
Consonants <f, v, w>                  & 10    & 90    & 10  & 100 & 30 & 100 \\ \hline
Consonant <g>                        & \textbf{30} & 100   & 10  & 100 & \textbf{80} & 100 \\ \hline
Consonants <g, ch>                   & 50    & 90    & 0   & 100 & 50 & 100 \\ \hline
Consonant <h>                        & \textbf{50}    & \textbf{70}    & 20  & 100 & 50 & 100 \\ \hline
Consonants <j, sch>                  & 20    & 100   & 0   & 100 & 20 & 100 \\ \hline
Consonants <k, x>                     & 50    & 100   & 10  & 100 & 40 & 100 \\ \hline
Consonant <s>                        &\textbf{ 80}    & 90    & 10  & 100 & \textbf{40} & 100 \\ \hline
Consonant <z>                        & 30    & 100   & 0   & 100 & 40 & 100 \\ \hline
n-Rule                     & 40    & 90    & 20  & 100 & 40 & 90  \\ \hline
French Loanwords              & 50    & 90    & 20  & 100 & 60 & 90  \\ \hline
Silent <e>                 & 20    & 90    & 0   & 100 & 30 & 100 \\ \hline
Plural French Loanwords          & \textbf{10}    & 100   & 0   & 100 & \textbf{10} & 100 \\ \hline
\end{tabular}
}
\caption{Success rate of Performance Tests, all scores are in percentages. The \textbf{correct} columns refer to sentences, where a correction is necessary, the \textbf{preserve} columns to sentences that should not be corrected. Results in bold are discussed in Section \ref{sec:res_qual}.}
\label{tab:behav_tests}
\end{table*}

\subsection{Qualitative} \label{sec:res_qual}
In a second step, we evaluate ByT5 base, mT5 and the \textit{spellux} pipeline qualitatively, focusing on models that are specifically trained for Luxembourgish, to compare the linguistic performance of each approach: byte-based, word-based and pipeline-based. Table \ref{tab:behav_tests} shows the results of this evaluation, with the scores indicating the success rate for the first (\textit{correct} columns) and second (\textit{preserve} columns) test setup. As described in Section \ref{sec:evaluation}, the first setup tests the correction of target words and the second setup the handling of false positives. 

Although ByT5 and \textit{spellux} have the same score in 7 categories, ByT5 performs better in 9 categories. In comparison, \textit{spellux} only performs better than ByT5 in 5 categories. The starkest differences in performance are present in the category <s> and <g>. ByT5 has a success rate of 80\% in comparison to the 40\% of \textit{spellux} in the <s> category. This category describes the orthographic rule for the unvoiced and voiced <s>, a phenomenon also influenced by the orthography of a related German word. Instead of correcting the incorrect form, \textit{spellux} keeps the input form, creating a false negative, or changing the word into a different word with a different meaning. For instance, in the test sentence with the target word \textit{iesen}, where the correct form would be \textit{iessen} (`to eat'), it corrects the word to \textit{eisen} (`ours'). On the other hand, \textit{spellux} has a higher success rate in the category <g> with 80\% compared to ByT5 with 30\%. This category describes the difference between the realization of <g> as a plosive and as a fricative \cite{Gilles_Trouvain_2013}. When the grapheme is realized as a fricative, the <g> is never doubled, as opposed to when the <g> is realized as a plosive. When looking into the output of ByT5, it can be seen that ByT5 keeps the target word the same instead of correcting it, creating a false negative. 

Interestingly, both the ByT5 and \textit{spellux} show the same low success rate of 10\% with the plural of French loanwords. French and German are both contact languages to Luxembourgish, which allowed for a rich borrowing history from both languages \cite{conrad_sprachkontakt}. This resulted in the orthographic inclusion of those words, particularly for French loanwords. Morphologically, the plural forms in Luxembourgish (<-en, -er>) are applied to French loanwords instead of French plural forms (<-s>). Due to the phonological phenomenon of deleting the <-n> ending before specific consonants, the <-e(n)> is replaced with <-ë> to avoid ambiguity \cite{peter_codification}. This rule is limited to French loanwords and both ByT5 and \textit{spellux} have a very low score. However, considering that the training data (the Chamber texts) contain many French loanwords – they are frequent in the political domain – it is somewhat surprising that ByT5 does not perform better in this category and \textit{spellux} might have achieved better results using the advanced correction modes. 

While ByT5 and \textit{spellux} perform similarly, mT5 shows low scores in every category. This aligns with our expectations based on the low performance in the quantitative evaluation. The best score for mT5 is 40\% in the neutral short vowel [\textipa{@}] category, a frequently realized sound in Luxembourgish \cite{Gilles_Trouvain_2013}. This is the written equivalent of the schwa, which is <e> for an unstressed syllable or <ë> for a stressed syllable.  

Overall, the second test setup (\textit{preserve} columns) indicates near perfect scores for all three models. Only ByT5 has a lower score of 70\% for the category <h> (vowel lengthening through <h> insertion) which is not used in written Luxembourgish but common in German. The lower performance in this category might be explained by the pre-training of the ByT5 model on different languages, including German. It is possible that false transfer learning from German to Luxembourgish could cause a lower performance. 

\section{Discussion}\label{sec:disc}
Automatic text normalization is a challenging task whose success depends on a number of factors, including a societally-anchored orthographic norm as the target of the correction task, the availability of large and standard-adherent datasets, suitable technical approaches for implementing the task, and a thorough understanding of the respective strengths and weaknesses of each approach. Given the current situation of written Luxembourgish – with a standard under development and limited amounts of correctly spelled text – in this paper we investigate text normalization approaches and present a comprehensive evaluation suite.
 
One of the challenges in developing text normalization tools is the use of synthetic data. While these are easy to produce based on existing corpora and orthographic rule sets, they do not represent the variation that would occur in real texts. To overcome these shortcomings, we present a new approach to generating training data with real-life variation data derived from actual texts written and corrected by writers of Luxembourgish. This approach has a clear advantage over synthetic data or prompting LLMs, as it represents the variation space of a language realistically, according to the actual writing practices of its speakers. In particular, the combination of variants and their frequency of use allows the creation of training data that reflect the variation patterns found in real-life texts.
 
Another problem with automatic text normalization is model evaluation. Given the large amount of variation found in written Luxembourgish, our approach to model evaluation includes a comprehensive set of quantitative and qualitative tests. These allow for a more fine-grained and linguistically informed analysis of the model output, e.g. by comparing success rates for specific orthographic rules. In this way, our evaluation suite increases the transparency of traditional evaluation metrics.
 
The results of the evaluation experiments show that the tested approaches not only perform differently in terms of quantitative success, e.g. ERR, but also show particular strengths and weaknesses for specific orthographic rules and contextual phenomena. In general, the latest version of GPT (October 2024) outperforms all other approaches, both model-based and pipeline-based. At the same time, the ByT5 model presented in this paper and the \textit{spellux} correction pipeline show individual strengths for certain sets of orthographic phenomena. Nevertheless, we believe that working with a technical solution tailored to Luxembourgish can be advantageous. First, our approach allows us to control all aspects of model training, i.e., training data, model parameters, and task implementation. Second, the use of real-life variation data as a basis for model training brings our approach closer to the actual variation space found in writing practice. Third, since the standard orthography is still under development, we can easily adapt and optimize our approach to future versions of the standard. Fourth, by combining linguistic analysis and hyperparameter optimization, our approach offers great potential for future iterations.
 
Looking beyond the task of text normalization, our approach can also serve as a linguistic analysis tool for detecting and classifying variation patterns in written Luxembourgish, for example in the context of the research project \textit{Tracing Attitudes And Variation In Online Luxembourgish Text Archives} (TRAVOLTA).\footnote{\url{https://www.uni.lu/fhse-en/research-projects/travolta/}} Using journalistic texts and user comments from the media platform RTL.lu, we can trace the development of individual as well as group-based writing practices outside the official spelling norm. Since there is hardly any research on the development of the written domain in Luxembourgish, the project can contribute to a better understanding of individual writing practices as well as the structure and dynamics of its variation space in general.

\section{Conclusion} 
In this paper, we present the first generative normalization model for Luxembourgish by creating training data from real-life variation data. More importantly, we develop performance tests for this normalizer to achieve a comprehensive, linguistically-informed evaluation using both quantitative and qualitative metrics. For the creation of training data, we use a variant dictionary with frequency information to create parallel training data with incorrect and correct sentence pairs. This training data is then used to fine-tune a ByT5 model and a mT5 model: the first sequence-to-sequence models fine-tuned for this task. Additionally, benchmarking is performed to compare byte-based (ByT5), word-based (mT5), LLM-based (Llama, GPT) and pipeline-based (\textit{spellux}) approaches. Furthermore, performance tests for Luxembourgish text normalization offer a deeper insight into the strengths and weaknesses of the models, as we compare ByT5, mT5 and \textit{spellux}. 

As the performance of ByT5 shows, our approach to the generation of training data is an effective method to train models while preserving a realistic variation space in the data. Furthermore, the ByT5 base model achieves comparable performances to other approaches with an ERR of 0.26. Overall, this paper shows that normalization for Luxembourgish is possible and achieves good results, either with prompting LLMs, using an already established pipeline, or with a ByT5 architecture. 

\section*{Limitations}
Due to the lack of a full-fledged standard in Luxembourgish, there is a very broad variation space with overlapping spelling variants. Therefore, the Spellchecker.lu variants should not be taken to reflect all possible variants in the variation space in Luxembourgish as they only reflect the users of the website.  

We have limited computing resources concerning specifically GPU space which results in a limited hyperparameter optimization setup. The GPU nodes available and used for the experiments are Dual CPU with 4 Nvidia accelerators and 768 GB RAM. 
 
\section*{Acknowledgments}
This research was supported by the Luxembourg National Research Fund (Project code: C22/SC/117225699).

We would also like to thank Emilia Milano and Verena Blaschke for their invaluable input, Jacques Spedener for the illustration, Lou Pepin for the Luxembourgish corrections, and Noah Lee for providing the idea for the Needleman-Wunsch algorithm.

\bibliography{custom}
\appendix
\section{Performance Test Units}

\begin{table*}[hbp]
\begin{adjustbox}{max width=\textwidth, center}
\begin{tabular}{@{}lll@{}}

\toprule
\multicolumn{2}{l}{Category} & Test Sentence \\ \midrule
\multicolumn{2}{l}{\bf Quantity rule}\\
\hspace{0.2em}
\punit
{writing of long vowels depending on stressed vowels \& consonants}
{\cite{peter_codification}}
{}
{Wou ass d'\underline{Bischt} fir ze kieren?}
{\textit{Correction:} Biischt}
{Where is the broom to sweep (with)?}

\multicolumn{2}{l}{\bf Short Vowels}\\
\hspace{0.2em}
\punit
{stressed short vowels and consonants}
{\cite{peter_luxembourgish}}
{}
{D'Haus ass op mech \underline{geschriwen}.}
{\textit{Correction:} geschriwwen}
{The house is written under my name.}

\multicolumn{2}{l}{\bf Short Open Vowel [\textipa{\ae}]}\\
\hspace{0.2em}
\punit
{distinction between <e> and <ä>}
{\cite{peter_codification}}
{}
{D'\underline{Mässere} si frësch geschlaff!}
{\textit{Correction:} Messere}
{The knives have been sharpened.}

\multicolumn{2}{l}{\bf Short Closed Vowel [\textipa{e}]}\\
\hspace{0.2em}
\punit
{distinction between <e> and <é>}
{\cite{peter_codification}}
{}
{Meng \underline{Wunnéng} ass um drëtte Stack.}
{\textit{Correction:} Wunneng}
{My flat is on the third floor.}

\multicolumn{2}{l}{\bf Neutral Short Vowel [\textipa{@}]}\\
\hspace{0.2em}
\punit
{distinction between <e> and <ë> for schwa sound}
{\cite{peter_phonolocical_domains}}
{}
{Kämm der deng \underline{Hoër}!}
{\textit{Correction:} Hoer}
{Comb your hair!}

\multicolumn{2}{l}{\bf Long Vowel [\textipa{e:}]}\\
\hspace{0.2em}
\punit
{distinction between <e> and <ee>}
{\cite{peter_codification}}
{}
{Mäi beschte Frënd ass \underline{Chines}.}
{\textit{Correction:} Chinees}
{My best friend is chinese.}

\multicolumn{2}{l}{\bf Diphthongs}\\
\hspace{0.2em}
\punit
{distinction between the 8 diphthongs}
{\cite{Gilles_Trouvain_2013}}
{}
{Firwat hues de dat net \underline{gleich} gesot?}
{\textit{Correction:} gläich}
{Why didn't you say that straight away?}

\multicolumn{2}{l}{\bf r-Rule}\\
\hspace{0.2em}
\punit
{distinction between consonant <r> and vocalized}
{\cite{peter_codification}}
{}
{De Poulet ass nach net ganz \underline{durch}.}
{\textit{Correction:} duerch}
{The chicken is not quite done yet.}

\multicolumn{2}{l}{\bf Final Devoicing}\\
\hspace{0.2em}
\punit
{distinction of voiced and unvoiced final consonants}
{\cite{peter_plosive}}
{}
{Eise Projet huet eng \underline{zolitt} Basis.}
{\textit{Correction:} zolidd}
{Our project has a solid base.}

\multicolumn{2}{l}{\bf Consonants <f, v, w>}\\
\hspace{0.2em}
\punit
{distinction between <f, v, w> based on German}
{\cite{peter_codification}}
{}
{Du waars e \underline{brawe} Jong.}
{\textit{Correction:} brave}
{You were a good boy.}

\multicolumn{2}{l}{\bf Consonant <g>}\\
\hspace{0.2em}
\punit
{distinction between <g> as a plosive and fricative}
{\cite{VariationdurchSprachkontakt}}
{}
{Hues du mech op dëser Foto \underline{getagt}?}
{\textit{Correction:}  getaggt}
{Did you tag me on this photo?}

\bottomrule
\end{tabular}%
\end{adjustbox}
\caption{Performance test units (part 1).}
\label{tab:performance-app1}
\end{table*}
\renewcommand{\arraystretch}{1}

\begin{table*}[hbp]
\begin{adjustbox}{max width=\textwidth, center}
\begin{tabular}{@{}lll@{}}

\toprule
\multicolumn{2}{l}{Category} & Test Sentence \\ \midrule
\multicolumn{2}{l}{\bf Consonants <g, ch>}\\
\hspace{0.2em}
\punit
{distinction between writings of fricatives after vowels}
{\cite{peter_codification}}
{}
{Ech wunnen an der \underline{Buerch}.}
{\textit{Correction:} Buerg}
{I live next to the castle.}

\multicolumn{2}{l}{\bf Consonant <h>}\\
\hspace{0.2em}
\punit
{consonant <h> and non-existent expansion <h>}
{}
{}
{All eis \underline{Méih} war ëmsoss!}
{\textit{Correction:} Méi}
{All our effort was for nothing.}

\multicolumn{2}{l}{\bf Consonants <j, sch>}\\
\hspace{0.2em}
\punit
{writing of fricatives}
{\cite{VariationdurchSprachkontakt}}
{}
{Am Zuch hunn e puer Leit Kaart \underline{geschpillt}.}
{\textit{Correction:} gespillt}
{A few people were playing cards on the train.}

\multicolumn{2}{l}{\bf Consonants <k, x>}\\
\hspace{0.2em}
\punit
{writing of consonants <k,x>}
{}
{}
{Dat Kand huet e gudde \underline{Karakter}.}
{\textit{Correction:} Charakter}
{This child has a good character.}

\multicolumn{2}{l}{\bf Consonant <s>}\\
\hspace{0.2em}
\punit
{distinction of voiced and unvoiced <s>}
{\cite{peter_codification}}
{}
{Mir \underline{iesen} de Mëtteg Nuddelen.}
{\textit{Correction:} iessen }
{We're having pasta for lunch.}

\multicolumn{2}{l}{\bf Consonant <z>}\\
\hspace{0.2em}
\punit
{distinction between <z> and <tz>}
{}
{}
{Hie geréit fënnef Keele mat enger \underline{Klaz}.}
{\textit{Correction:} Klatz}
{He knocked down 5 pins with one ball.}

\multicolumn{2}{l}{\bf n-Rule}\\
\hspace{0.2em}
\punit
{deletion of final <-n> before specific characters}
{\cite{peter_nregel}}
{}
{De Theo war am Orall op Zak.}
{\textit{Correction:} De}
{Thea was quick to answer in his oral exam.}

\multicolumn{2}{l}{\bf French Loanwords}\\
\hspace{0.2em}
\punit
{writing of French loanwords}
{\cite{conrad_sprachkontakt}}
{}
{Hues du deng \underline{Valise} scho gepaakt?}
{\textit{Correction:} Wallis}
{Have you packed your case already?}

\multicolumn{2}{l}{\bf Silent <e>}\\
\hspace{0.2em}
\punit
{silent <e> of French loanwords}
{\cite{peter_phonolocical_domains}}
{}
{Ech ginn ni ouni \underline{Necessair} op d'Rees.}
{\textit{Correction:} Necessaire}
{I will never go without my sewing kit on vacation.}

\multicolumn{2}{l}{\bf Plural French Loanwords}\\
\hspace{0.2em}
\punit
{plural of French loanwords <-er, -en, -ë, -éen, -éë>}
{\cite{conrad_sprachkontakt}}
{}
{Mir kréien am Fréijoer nei \underline{Faccen}.}
{\textit{Correction:} Facen}
{We are getting a new facade in spring.}

\bottomrule
\end{tabular}%
\end{adjustbox}
\caption{Performance test units (part 2).}
\label{tab:performance-app2}
\end{table*}
\renewcommand{\arraystretch}{1}


\end{document}